\documentclass{article}
\usepackage{spconf,amsmath,graphicx}
\usepackage[hidelinks]{hyperref}
\usepackage{amsmath}
\usepackage{amssymb}
\usepackage{mathtools}
\usepackage{amsthm}
\usepackage{multirow}
\usepackage{makecell}
\usepackage{booktabs}
\usepackage{pifont}
\usepackage{tabularx}
\usepackage{enumitem}
 \usepackage{cancel}
 \usepackage{dsfont}
\usepackage[table]{xcolor}
\newcommand{\shu}{\textcolor{red}}
\newcommand{\dat}{\textcolor{blue}}
\hypersetup{
    colorlinks=true,
    linkcolor=blue,
    filecolor=magenta,      
    urlcolor=cyan,
    citecolor=blue,
    citecolor=blue,
}

\newcommand{\mathbbm}[1]{\mathds{#1}}
\title{Towards Multi-Modal Forgery Representation Learning for AI-Generated Video Detection and Localization}
\name{Dat Le$^{1 \star}$ \thanks{$\star$Co-first Authors} \qquad Khoa Nguyen$^{1 \star}$ 
\qquad Xin Wang$^{2}$
\qquad Shu Hu$^{1 \dagger}$\thanks{$\dagger$Corresponding Author, {\tt hu968@purdue.edu}. }}

\address{$^1$ Purdue University, West Lafayette, {\tt \small \{le317, nguy1053, hu968\}@purdue.edu} \\
$^2$University at Albany, SUNY, New York, {\tt \small xwang56@albany.edu}}
\begin{document}
%\ninept
%
\maketitle
\begin{abstract}
Recent advances in generative AI have democratized video creation at scale. AI-generated videos, including partially manipulated clips across visual and audio channels, pose escalating risks of semantic distortion and misuse, which motivates the need for reliable detection tools. Most existing AI-generated video detectors remain limited by single- or partial-modality of data modeling and the lack of fine-grained temporal forgery localization. To address these challenges, our primary novelty introduces a core architecture that jointly integrates an LMM semantic branch with a spatio-temporal (ST) visual branch and a multi-scale partial-spoof (PS) audio branch. This multi-modal approach enables simultaneous detection and fine-grained temporal localization of partially manipulated AI-generated video forgeries. Extensive experiments show that this approach outperforms existing state-of-the-art methods.
The code is available at \dat{\textit{\url{https://github.com/Purdue-M2/Deepfake-Detection}}}.
\end{abstract}
\vspace{-2mm}
\begin{keywords}
AI-generated Video, Large Language Models, Deepfake Detection, Localization 
\end{keywords}
\vspace{-3mm}
\section{Introduction}
\label{sec:intro}
\vspace{-2mm}
Recent progress in generative AI  has made video creation dramatically cheaper and faster by automating production steps, enabling scalable personalization, 
% (e.g., tailored advertising and training content), 
and lowering the barrier to entry for non-experts. At the same time, modern systems are no longer limited to silent footage: recent text-to-video models can generate videos with \emph{native audio}, including synchronized sound and speech, and are being deployed through widely accessible platforms and APIs (e.g., Veo 3.1 via Gemini \cite{team2023gemini}).
This rapid democratization expands creative opportunities, but it also increases the risk and scale of misuse.

AI-generated videos can be manipulated at multiple levels, most notably the \emph{visual} and \emph{audio} channels. Beyond fully synthetic clips, an increasingly concerning threat is \emph{partial} manipulation, where only short segments are altered while the rest of the content remains authentic. Such targeted edits can flip the semantic meaning of an otherwise real recording, 
% (e.g., changing ``I don't like ice cream'' to ``I like ice cream''), 
which raises high-stakes concerns in domains such as politics, journalism, and social media. This threat is not hypothetical: industry and policy reports highlight deepfakes as a fast-growing driver of fraud, impersonation, and misinformation, motivating the need for reliable detection tools.
\begin{figure}[]
    \centering
    \includegraphics[width=0.8\linewidth]{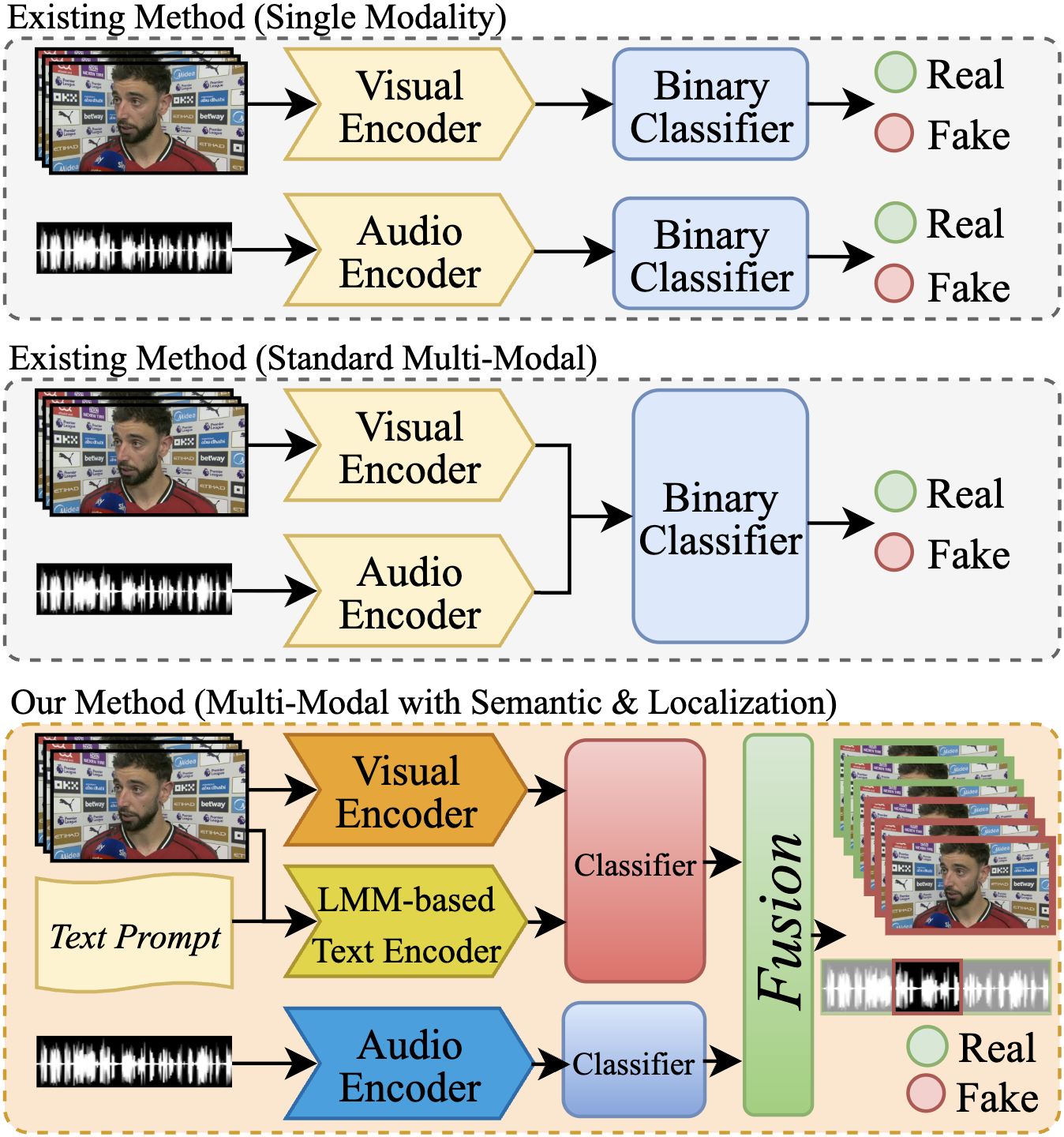}
    \vspace{-5mm}
    \caption{\small \textit{Comparison between existing methods and ours. The top and middle rows illustrate existing single-modality and multi-modal methods that perform binary classification. The bottom row depicts our  framework, which integrates an LMM-based semantic branch to enable fine-grained temporal localization beyond the detection}.}
    \label{fig:intro}
    \vspace{-0.5cm}
\end{figure}

Most existing deepfake detection methods \cite{lin2024detecting,lin2024preserving} still exhibit at least one of the following limitations. 
(i) Most of them focus on a single modality (visual-only or audio-only) without explicitly leveraging complementary signals \textbf{across modalities}. 
{(ii) Some \cite{Zhang_2023} consider only two modalities (typically audio-visual) while \textbf{overlooking semantic and language-driven cues} that can expose semantic-level inconsistencies.}
(iii) A few consider language-driven cues \cite{song2025learningmultimodalforgeryrepresentation}, but they treat detection as a coarse binary classification problem, assuming the entire clip is either real or fake, and \textbf{do not provide fine-grained temporal forgery localization}. 

% \begin{table}[t]
% \centering
% \caption{Existing works by modality and localization ability.}
% \label{tab:related_works_modalities}
% \small
% \setlength{\tabcolsep}{3.5pt}
% \renewcommand{\arraystretch}{1.1}
% \scalebox{0.8}{
% \begin{tabularx}{\columnwidth}{@{}Xcccc@{}}
% \toprule
% \textbf{Method} & \textbf{Text} & \textbf{Visual} & \textbf{Audio} & \textbf{Loc.} \\
% \midrule
% PyAnnote (Zero-Shot) \cite{Plaquet_2023} & \xmark & \xmark & \cmark & \cmark \\
% PartialSpoof \cite{10003971}        & \xmark & \xmark & \cmark & \cmark \\
% \midrule
% ActionFormer + InternVideo \cite{zhang2022actionformerlocalizingmomentsactions}, \cite{wang2022internvideogeneralvideofoundation}  & \xmark & \cmark & \xmark & \cmark \\
% ActionFormer + VideoMAEv2 \cite{zhang2022actionformerlocalizingmomentsactions} \cite{wang2023videomaev2scalingvideo} & \xmark & \cmark & \xmark & \cmark \\
% \midrule
% BA-TFD \cite{cai2023reallymeanthatcontent}    & \xmark & \cmark & \cmark & \cmark \\
% BA-TFD+ \cite{cai2023glitchmatrixlargescale}   & \xmark & \cmark & \cmark & \cmark \\
% UMMAFormer \cite{Zhang_2023} & \xmark & \cmark & \cmark & \cmark \\
% \midrule
% MM-Det \cite{song2025learningmultimodalforgeryrepresentation}& \cmark & \cmark & \xmark & \xmark \\
% \midrule
% \textbf{\textit{Ours}}  & \cmark & \cmark & \cmark & \cmark \\
% \bottomrule
% \end{tabularx}
% }
% \end{table}

% However, recent benchmarks and challenge settings emphasize that temporal localization is crucial because real-world forgeries are often embedded as short manipulated segments rather than full-length edits.

\begin{figure*}[!t]
    \centering
    \includegraphics[width=0.9\textwidth]{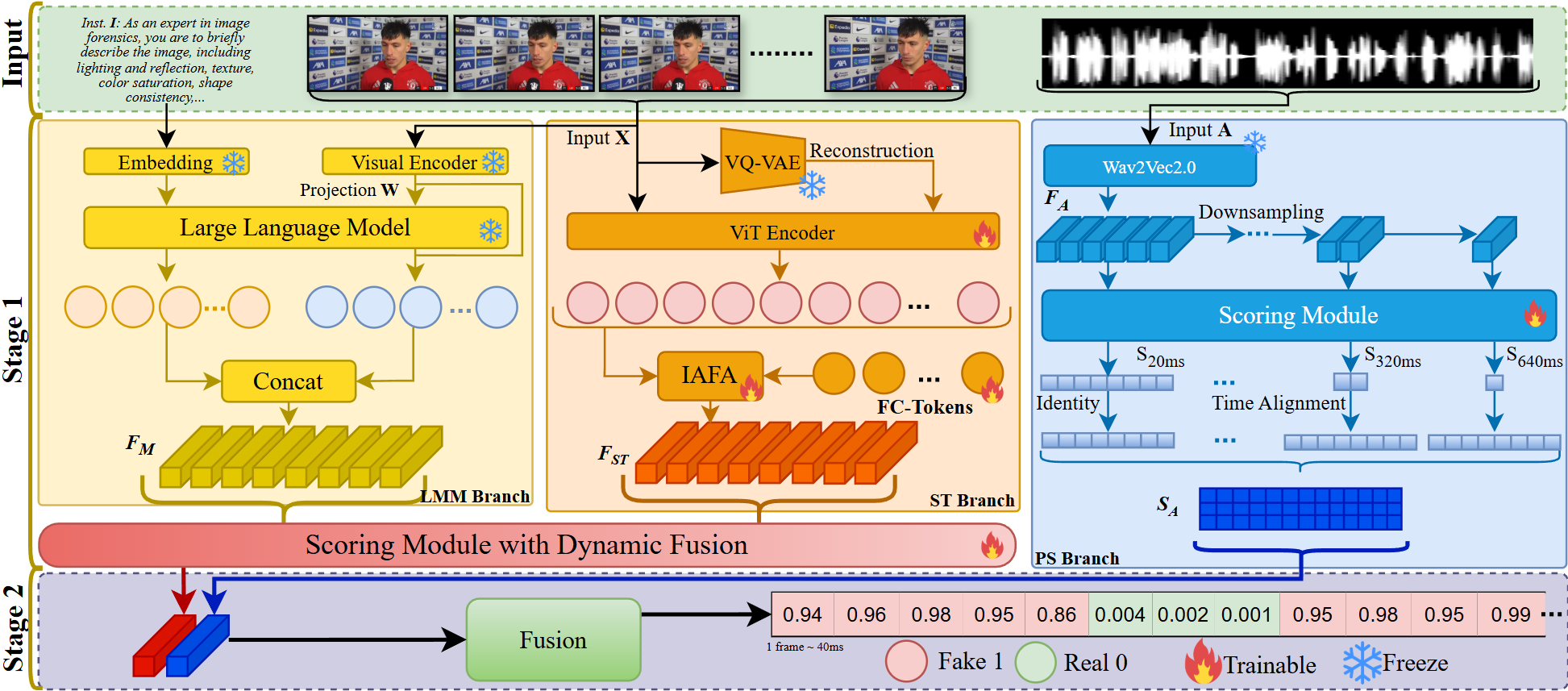}
    \vspace{-5mm}
    \caption{\small \textit{Overview of the proposed multi-branch framework for AI-generated video detection and localization. Visual content is processed by an LMM branch and an ST branch, while audio is analyzed by a PS branch that produces multi-resolution segment scores aligned to a common timeline. The branch outputs are integrated by a dynamic fusion module to generate dense per-segment predictions
    % (e.g., 40 ms stride). Snowflake and flame icons denote frozen and trainable components, respectively
    }.}
    \vspace{-0.5cm}
    \label{fig:main}
\end{figure*}

In this work, we propose a multi-branch framework that models deepfake evidence at three complementary levels: 
(i) \textbf{visual artifact} cues capturing pixel-level and temporal inconsistencies, 
(ii) \textbf{audio} cues capturing spoofing traces over multiple temporal resolutions, and 
(iii) \textbf{text/semantic} cues extracted via a \textit{Large Multi-modal Model} (LMM) that summarizes high-level inconsistencies and provides interpretable signals. 
These modality-specific prediction streams are then fused to produce temporally localized forgery likelihoods, enabling not only detection but also localization and modality-level attribution of the suspicious segments. Fig. \ref{fig:intro} shows the difference between the existing methods and ours.  The main contributions of this paper are as follows:
\begin{enumerate}[nosep]
    \item We introduce a novel core architecture that jointly integrates an LMM semantic branch with a spatio-temporal (ST) visual branch and a multi-scale (PS) audio branch.
    \item Beyond detection, this joint integration allows us to simultaneously achieve fine-grained temporal localization of partial manipulations by producing dense, time-aligned forgery likelihood streams.
    \item We demonstrate the effectiveness of the proposed approach through experiments that surpass state-of-the-art methods.
\end{enumerate}

% \section{Related Work}
% \label{sec:rw}

% \subsection{Video Generation by AI}
% \label{sec:rw_gen}

% \subsection{AI-generated Video Detection}

\vspace{-3mm}
\section{Method}

% Main
\vspace{-2mm}
% \subsection{Framework Overview} 
Our proposed framework (see Fig. \ref{fig:main}) is designed to detect and localize deepfakes via a two-stage multi-modal approach. 
% The framework integrates three distinct modalities: text, visual, and audio features. The pipeline operates in two primary stages: \textit{Forgery Exposure} and \textit{Detection and Localization}. 
% \begin{enumerate}
%     \item Forgery Exposure: Extracting features and generating independent forgery probabilities for each modality in fine grain segment
%     \item Detection and Localization: Aggregating these independent streams to localize forgeries over time.
% \end{enumerate}
\vspace{-3mm}
\subsection{Stage 1: Forgery Exposure}
% Given an input video $V$, we decompose it into a sequence of frames
% $X=\{x_i\}_{i=1}^{T}$ and an audio track $A$.
% We additionally provide a configurable instruction $I$, which can be replaced with alternative instruction variants.
% We extract the features for each modality as follows: 

\textbf{Exposing Multi-Modal Forgery via LMM}.
\label{sec:lmm_branch}
To obtain a more generalizable multi-modal forgery representation for open-world AI-generated videos, we adopt an LMM-based feature extraction branch following the LLaVA paradigm \cite{song2025learningmultimodalforgeryrepresentation, liu2023visualinstructiontuning}. Given a video $X=\{x_i\}_{i=1}^{T}$, each frame is encoded by a CLIP \cite{radford2021learningtransferablevisualmodels} visual encoder $\mathcal{D}_v$ to obtain a visual representation $F_V$.
In parallel, we define a fixed forensic instruction prompt $I$: ``As an expert in image forensics, you are to briefly describe the image, including lighting and reflection, texture, color saturation, shape consistency, sense of depth, compression trace, artifacts. Give a reason to justify whether it is a real or a fake image.'' adopted from \cite{song2025learningmultimodalforgeryrepresentation}. We process $I$ using the tokenizer and embedding layer of the pre-trained Large Language Model (LLM) to obtain the textual instruction embeddings, denoted as $F_I$.

To align the visual and textual modalities, the visual representation $F_V$ is passed through a trainable projection layer $\mathcal{W}$ (e.g., a linear projection or MLP) to map the visual features into the same embedding space as the text tokens. The aligned visual tokens and the instruction embeddings $F_I$ are then fed into the LLM decoder $\mathbb{D}$ to produce enhanced multi-modal hidden states: $F_L = \mathbb{D}(F_I, \mathcal{W}(F_V))$.
% \begin{equation}
% F_L = \mathbb{D}(F_I, \mathcal{W}(F_V)).
% \end{equation}

Finally, we construct the LMM feature by concatenating the projected visual and language hidden representations: $F_{\text{M}} = \mathrm{CONCAT}\big(\mathrm{PROJ}(F_V), \mathrm{PROJ}(F_L)\big),$
where $\mathrm{PROJ}(\cdot)$ denotes a linear projection used to align dimensions. This branch captures semantic inconsistencies and model-level reasoning signals complementary to pixel-level artifact features.
% \begin{equation}
% F_L = \mathbb{D}(F_I, \mathcal{W}(F_V)).
% \end{equation}

\smallskip
\noindent
\textbf{Exposing Spatio-Temporal Visual Forgery}.
To capture low-level diffusion artifacts and temporal inconsistencies, we adopt a reconstruction-guided spatio-temporal (\textbf{ST}) branch. We first reconstruct each frame using a VQ-VAE \cite{van2017neural} to obtain $\hat{x}_i$. The reconstruction amplifies generative traces, and the discrepancy between $x_i$ and $\hat{x}_i$ provides a strong cue for distinguishing real versus synthetic content. We therefore jointly feed both the original and reconstructed frames into a spatio-temporal encoder: $F_{ST} = \mathcal{E}_{ST}(X,\hat{X}),$
% \begin{equation}
%     F_{ST} = \mathcal{E}_{ST}(X,\hat{X}),   
% \end{equation}
where $\hat{X}=\{\hat{x}_i\}_{i=1}^T$ and  $\mathcal{E}_{ST}$ is a Hybrid ViT equipped with In-and-Across Frame Attention (IAFA) to effectively aggregate local and global forgery patterns.

Specifically, IAFA utilizes two distinct types of tokens: Patch tokens (P-tokens) representing local spatial regions, and Frame-Centric tokens (FC-tokens) designed to encapsulate the global authenticity information of each frame. Within each Transformer block, IAFA performs two alternating attention operations: (i) \textit{Global Patch Attention}: P-tokens compute self-attention across all P-tokens from all video frames. This allows the model to capture fine-grained spatial artifacts and track temporal inconsistencies across the entire video sequence. (ii) \textit{Frame-Centric Aggregation}: Subsequently, the FC-token for each frame attends to all P-tokens within that same frame. This operation aggregates the distributed local forgery clues into a unified frame-level representation, enabling the model to summarize global forgery evidence from local anomalies. The features from the ST and LMM branches are then processed by a Scoring Module with Dynamic Fusion, which employs a multi-head attention mechanism, to obtain the final forgery prediction score $S_V$.

\smallskip
\noindent
\textbf{Exposing Audio Forgery}.
To capture synthetic speech artifacts across varying temporal scales, we adopt the PartialSpoof Countermeasure (CM) architecture \cite{10003971} to design a PS branch. Given an input waveform $A$, a pre-trained SSL front-end (Wav2Vec2.0 \cite{NEURIPS2020_92d1e1eb}) extracts frame-level representations $\mathbf{a}_{1:N}$ at a $20\,\mathrm{ms}$ frame shift. The PartialSpoof back-end then computes segment-level spoofing scores at six temporal resolutions $\{20,40,80,160,320,640\}\,\mathrm{ms}$.

At the finest level ($20\,\mathrm{ms}$), scores are generated directly from the SSL features.
For coarser levels ($40$--$640\,\mathrm{ms}$), the feature sequence is progressively downsampled in a fine-to-coarse manner. 
At each resolution $r$, the resulting features are fed to a lightweight gMLP scoring head composed of five stacked gMLP blocks, following PartialSpoof \cite{10003971}. 
Here, gMLP is a gated MLP variant that augments a standard feed-forward MLP with a gating mechanism for sequence mixing, offering a simple yet effective alternative to heavier architectures \cite{DBLP:journals/corr/abs-2105-08050}. 
This yields a segment-level score sequence $\mathbf{s}_r$.

To match the temporal stride of the other branches in our fusion module (approximately one token per $40\,\mathrm{ms}$), we treat $40\,\mathrm{ms}$ as the finest score in our pipeline. Accordingly, we discard the $20\,\mathrm{ms}$ scores and align all remaining resolutions to a $40\,\mathrm{ms}$ grid for fusion. Since coarser resolutions yield fewer score tokens, we align scores to the $40\,\mathrm{ms}$ grid using repeat-based upsampling, where each score at resolution $r$ is repeated $\frac{r}{40\,\mathrm{ms}}$ times, resulting in an aligned score sequence $\mathbf{s}_{r}\in\mathbb{R}^{T_{40}}$, where $T_{40}$ denotes the total number of $40\,\mathrm{ms}$ frames in the audio clip (i.e., $\mathrm{len}(A) / 40\,\mathrm{ms}$).

\begin{table}[t]
\centering
\vspace{-3mm}
\caption{\small \textit{Segment-level performance of single-resolution audio scoring heads on a AV-Deepfake1M++ \cite{cai2025av} (audio only)}.}
\label{tab:audio_res_ablation}
\begin{tabular}{c|cc}
\hline
Resolution (ms) & AUC $\uparrow$ & EER $\downarrow$ \\
\hline
40  & 99.37 & 1.12 \\
80  & 99.39 & \textbf{1.10} \\
160 & 99.42 & 1.14 \\
320 & \textbf{99.71} & 1.11 \\
640 & 99.31 & 1.33 \\
\hline
\end{tabular}
\vspace{-0.5cm}
\end{table}
In preliminary fusion experiments, we found that retaining only the coarser resolutions $\{160,320,640\}\,\mathrm{ms}$ yields the best synergy with the other branches.
Specifically, Table~\ref{tab:audio_res_ablation} reports the segment-level performance of single-resolution audio heads on AV-Deepfake1M++ \cite{cai2025av}(audio only): mid/coarse windows perform consistently well, with $320\,\mathrm{ms}$ achieving the highest AUC.
We conjecture this is because $160$--$640\,\mathrm{ms}$ better matches the typical temporal extent of word-level acoustic events and the average fake-segment duration in our data, whereas very fine windows (e.g., $40$--$80\,\mathrm{ms}$) are more sensitive to local variability and add redundancy when fused with other branches.
We therefore discard finer scales and construct the final audio representation by stacking only these three aligned score sequences: $S_A =
\big[\mathbf{s}_{160}, \mathbf{s}_{320}, \mathbf{s}_{640}\big]
\in \mathbb{R}^{3 \times T_{40}}.$
% \begin{equation}
% S_A =
% \big[\mathbf{s}_{160}, \mathbf{s}_{320}, \mathbf{s}_{640}\big]
% \in \mathbb{R}^{3 \times T_{40}}.
% \end{equation}
This choice also reduces compute and memory overhead in our multi-branch pipeline, since we must store dense segment-level predictions across the full audio timeline for downstream processing.

\smallskip
\noindent
\textbf{Training Strategy}.
We train the visual and audio pipelines separately.
For the visual pipeline, we jointly optimize the ST branch and a Dynamic Fusion module. We optimize the trainable parameters $\theta_{V}$ (ST, LMM, and Dynamic Fusion) using the binary cross-entropy  \cite{song2025learningmultimodalforgeryrepresentation}: $\mathcal{L}_{V}(\theta_{V}) = \frac{1}{T}\sum_{i=1}^T\mathcal{C}\big(S_{V}^i, y_i  \big)$,
% \begin{equation}
% \mathcal{L}_{V}(\theta_{V}) = \frac{1}{T}\sum_{i=1}^T\mathcal{C}\big(S_{V}^i, y_i  \big),
% \end{equation}
where $S_V^i$ denotes the predicted score for sample $i$ and $y_i$ is its ground-truth. $\mathcal{C}(\cdot,\cdot)$ is the cross-entropy loss. 
% $[y_1, y_2]$ is the one-hot encoded frame-level ground-truth label, and $S_{V,1}$ and $S_{V,2}$ correspond to the probabilities of the two binary classes.

% \shu{rewrite this part. Please answer my questions: You mentioned dynamic fusion. However, you did not provide more details about this fusion. What is the difference between this dynamic fusion and the traditional feature fusion? If they are the same, why not directly say fusion? what is the relationship between $S_V$ and $v$? }

For the audio pipeline, we extract the audio track $A$ and pass it through a frozen SSL feature extractor (Wav2Vec2.0) to obtain $F_A$.
We train only the PS scoring module to predict the audio score $S_A$ using $10$\,ms ground-truth labels aligned to the scoring resolution.
We optimize the PS scoring module using the P2SGrad-based MSE loss \cite{10003971, wang2021comparativestudyrecentneural, DBLP:journals/corr/abs-1905-02479}: $\mathcal{L}_{A}
=
\frac{1}{|\mathcal{D}|}
\sum_{j=1}^{|\mathcal{D}|}
\sum_{k=1}^{C}
\Bigl(
\cos\theta_{j,k} - \mathbbm{1}(y_j = k)
\Bigr)^2$,
% \begin{equation}
% \mathcal{L}_{A}^{(\mathrm{P2S})}
% =
% \frac{1}{|\mathcal{D}|}
% \sum_{j=1}^{|\mathcal{D}|}
% \sum_{k=1}^{C}
% \Bigl(
% \cos\theta_{j,k} - \mathbbm{1}(y_j = k)
% \Bigr)^2,
% \end{equation}
where $\cos\theta_{j,k}$ denotes the cosine similarity between the sample embedding and class prototype (real vs.\ fake), and $\mathbbm{1}(\cdot)$ is the indicator function. $|\mathcal{D}|$ is the number of training samples and $C$ is the number of classes (set to $2$ for real vs.\ fake).

% \shu{add OnecycleLR}
We use a stochastic gradient descent-based approach to optimize each model using its corresponding loss. However, due to the large size of the dataset, the training loss typically plateaus during the later epochs and exhibits minor fluctuations around the plateau. To this end, we adopt the One-Cycle learning rate policy \cite{smith2019super}. The One-Cycle scheduler first increases the learning rate to a predefined maximum value and then gradually decreases it over the remaining training iterations. This strategy encourages faster convergence and helps the optimizer escape flat regions of the loss landscape. 
% In our experiments, One-Cycle learning rate scheduling leads to a more stable optimization process and achieves better performance compared to ReduceLROnPlateau.

\subsection{Stage 2: Detection and Localization} 

\textbf{Detection using Fusion Module}.
The fusion module operates on the logits produced by multiple sub-models. Specifically, two logits (for real and fake) are obtained from the ST branch and LMM branch after passing through the Scoring Module with Dynamic Fusion, and six logits are obtained from the multi-scale temporal models, with 2 logits each for temporal resolutions of 160, 320, and 640 milliseconds. These 8 logits are concatenated and passed through a linear fusion module, which outputs 2 final logits corresponding to real and fake predictions. A softmax function is applied to these two logits to obtain the frame-level class confidence scores. The model produces a sequence of frame-level confidence scores. As 0 denotes real and 1 denotes fake in the labels, the video-level prediction is obtained by selecting the maximum frame-level score across all frames.

% \shu{I remember you used softmax, right? Add this discussion}
\smallskip
\noindent
\textbf{Temporal Localization}.
We introduce a segment-based training and inference scheme to localize the fake region. During training, we divide each video into segments (sliding windows). A segment is labeled as fake if it overlaps with any ground-truth fake time interval, and real otherwise. 
\begin{enumerate}[nosep]
    \item \textbf{LMM branch.} The LMM branch produces sparse predictions every 8 seconds (200 frames). We densify the timeline via nearest-neighbor interpolation to obtain an aligned sequence $F_{M}$ for fusion (Fig. \ref{fig:lmm_localization}).
\begin{figure}[t]
    \centering
    \includegraphics[width=0.95\linewidth]{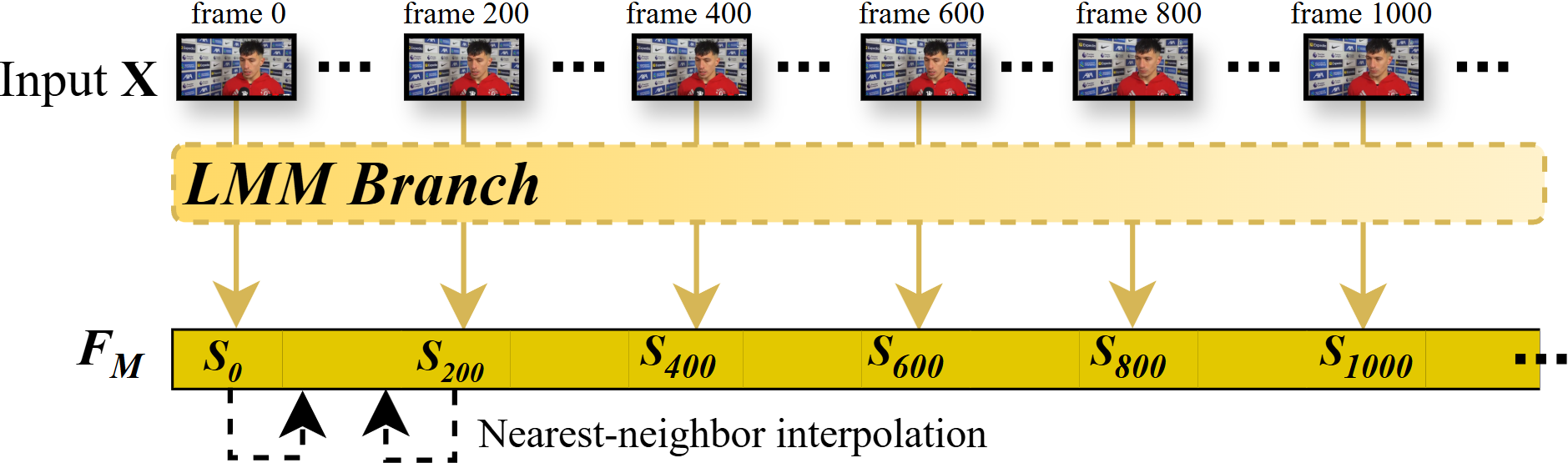}
    \vspace{-5mm}
    \caption{\small \textit{LMM-branch temporal localization via sparse sampling. Predictions are made every 8 seconds and the timeline $F_{M}$ is densified using nearest-neighbor interpolation.}}
    \vspace{-4mm}
    \label{fig:lmm_localization}
\end{figure}

\item \textbf{ST branch.}
For the ST branch, we adopt a sliding window prediction mechanism:
(i) a fixed-length window of frames is fed into the model;
(ii) the predicted score of the window is assigned to the center frame of that window;
(iii) the window is then shifted forward by one frame and repeated until the end of the video.
For boundary cases, the first few frames reuse the prediction from the first valid center frame, while the last few frames reuse the prediction from the last valid window. (see Fig. \ref{fig:st_localization})

% This strategy enables frame-level temporal localization while maintaining compatibility with window-based models (Fig. \ref{fig:st_localization}).
\begin{figure}[t]
    \centering
    \includegraphics[width=0.95\linewidth]{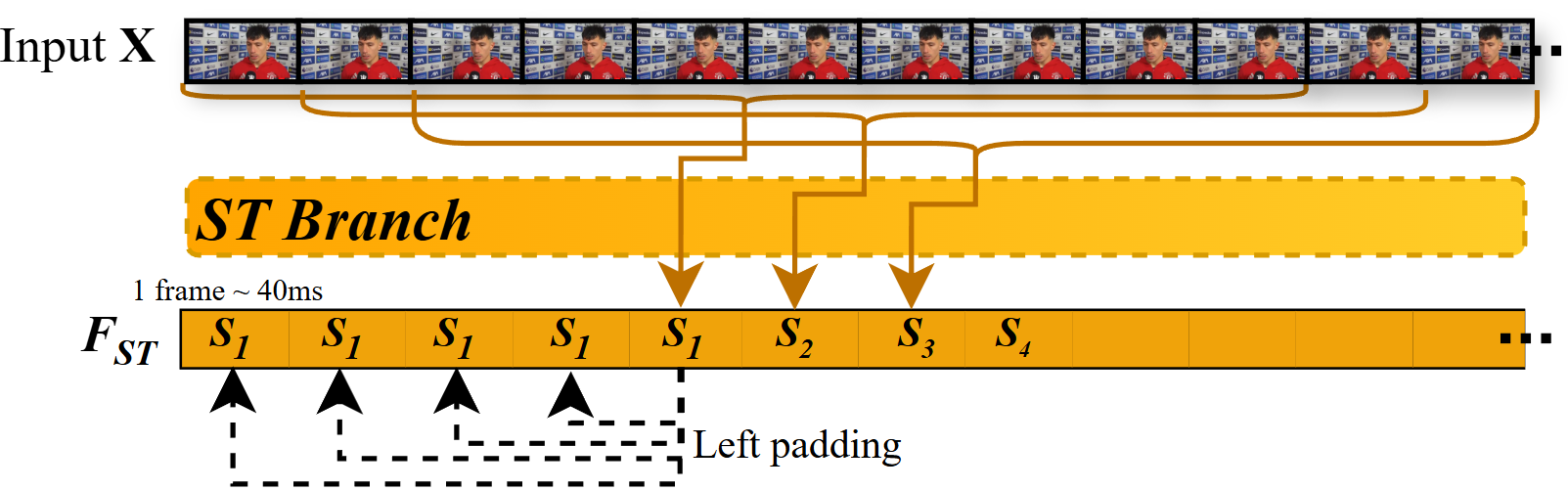}
    \vspace{-5mm}
    \caption{\small \textit{ST-branch temporal localization via a sliding window.
    % The video is scanned with a fixed-length window to form $F_{ST}$ at a $\sim$40\,ms stride; padding is applied at both the start and end for boundary handling
    }}
    \label{fig:st_localization}
    \vspace{-7mm}
\end{figure}

\item \textbf{PS branch.}
The PS result $F_{A}$ natively outputs fine-grained segment-level scores and therefore its output is directly used for localization.
\end{enumerate}

\section{EXPERIMENTS}

% ==========================================
% SETTINGS
% ==========================================
\vspace{-3mm}
\subsection{Settings}

\noindent \textbf{Datasets.} We primarily evaluate our framework on the {AV-Deepfake1M++}\cite{cai2025av} dataset, a challenging large-scale benchmark designed for audio-visual deepfake detection. This dataset is particularly rigorous due to its inclusion of extremely short manipulated segments, averaging approximately 320 ms. These brief perturbations require high-precision temporal localization and make global binary classification significantly more difficult compared to traditional datasets. Additionally, to evaluate the cross-dataset generalization capability of our model, we conduct zero-shot inference on the {FakeAVCeleb} \cite{khalid2022fakeavcelebnovelaudiovideomultimodal} dataset, using the model trained solely on AV-Deepfake1M++.

\smallskip
\noindent \textbf{Evaluation Metrics.} For video-level detection, we report the Area Under the Receiver Operating Characteristic Curve (\textbf{AUC (Video)}). For fine-grained temporal localization, we report the (1) segment-level AUC (\textbf{AUC (Seg)}), and (2) Average Precision (\textbf{AP}) and Average Recall (\textbf{AR}) at distinct Intersection over Union (IoU) thresholds (0.5, 0.75, 0.9, 0.95) and various proposal counts (5, 10, 20, 30, 50), respectively. 
% at various Intersection over Union thresholds (e.g., AP@0.5, AP@0.75), which strictly measure the overlap between predicted and ground-truth forgery intervals.

\smallskip
\noindent \textbf{Baselines.} We compare our method against three state-of-the-art baselines: \textit{MM-Det\cite{song2025learningmultimodalforgeryrepresentation} (Text+Visual):} A recent multi-modal baseline leveraging text and visual cues but lacking a dedicated audio spoofing branch. \textit{UMMAFormer\cite{Zhang_2023}, BA-TFD/BA-TFD+\cite{cai2022you} (Visual+Audio) :} transformer-based frameworks that model audio-visual interactions.

\smallskip
\noindent \textbf{Implementation Details.} Our model is trained with a multi-task objective combining binary classification and dense segment regression. We utilize the {OneCycleLR} scheduler to ensure stable convergence. The final predictions are obtained via a Softmax layer over the fused logits. The temporal resolution for localization is set to 40 ms per token.

% --- FIGURE: QUALITATIVE (Spanning 2 columns at the top) ---
\begin{figure*}[t!]
  \centering
  \includegraphics[width=0.9\linewidth]{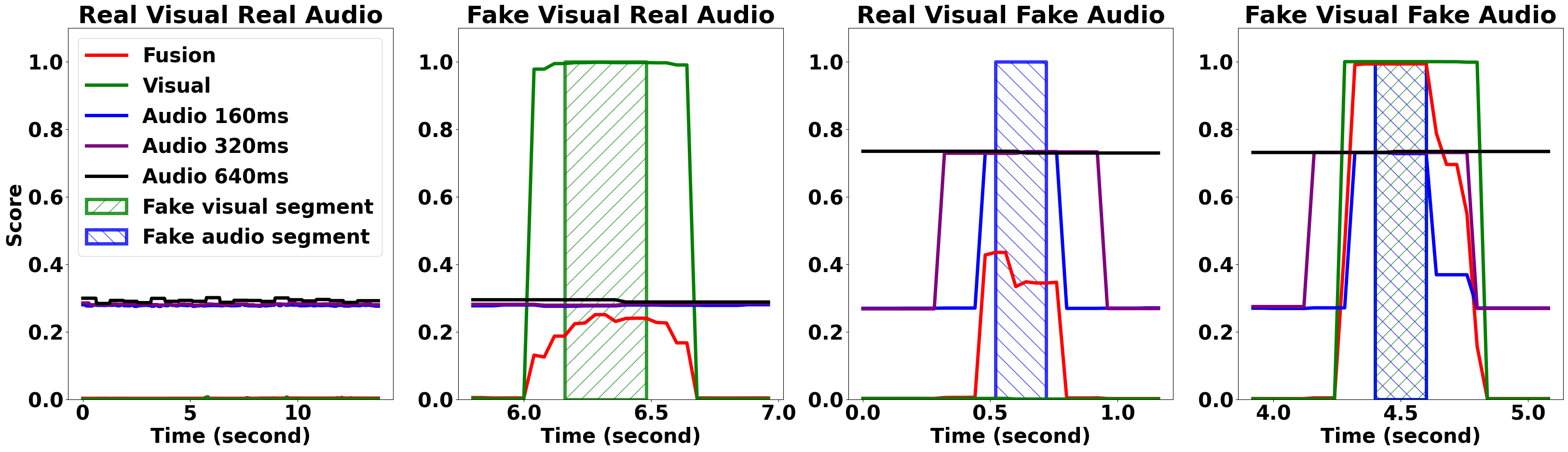} 
  \vspace{-5mm}
  % Note: added \linewidth to keep aspect ratio
  \caption{\small \textit{Visualization of temporal localization scores across different forgery scenarios in AV-Deepfake1M++.
  % From left to right: real video real audio, fake video real audio (Visual branch drives detection), real video fake audio (Audio branches drive detection), fake video fake audio. 
  The {Fusion} score (\shu{red}) robustly localizes the manipulated segments (shaded regions) even when one modality fails to detect a forgery.}}
  \vspace{-5mm}
  \label{fig:qualitative}
\end{figure*}
% ==========================================
% RESULTS (DETECTION & ABLATION)
% ==========================================
\vspace{-3mm}
\subsection{Quantitative Results}
\vspace{-2mm}
% --- TABLE: DETECTION MAIN (Placed strictly at start of Results) ---
\begin{table}[t]
\centering
\caption{\small \textit{Comparison of detection performance (\%)}.}
\label{tab:detection_main}
\resizebox{\linewidth}{!}{
\begin{tabular}{c|cc|c}
\hline
\multirow{2}{*}{\textbf{Method}} & \multicolumn{2}{c|}{\textbf{AV-Deepfake1M++}} & \textbf{FakeAVCeleb} \\
 & \textbf{AUC (Video)} $\uparrow$ & \textbf{AUC (Seg)} $\uparrow$ & \textbf{AUC (Video)} $\uparrow$ \\ \hline
BA-TFD (Visual+Audio) \cite{cai2022you} & 50.28 & 81.15 & 49.27\\
BA-TFD+ (Visual+Audio) \cite{cai2022you} & 55.43 & 82.19 & 51.70 \\
MM-Det (Text+Visual) \cite{song2025learningmultimodalforgeryrepresentation} & 82.61 & 80.83 & 76.74 \\
UMMAFormer (Visual+Audio) \cite{Zhang_2023} & 87.04 & 91.02 & 62.29 \\ \hline
\textbf{Ours} & \textbf{96.66} & \textbf{98.23} & \textbf{83.03} \\ \hline
\end{tabular}
}
\vspace{-0.5cm}
\end{table}

\textbf{Comparative Detection Performance.}
Table \ref{tab:detection_main} summarizes the detection performance on AV-Deepfake1M++ and FakeAVCeleb. Our proposed method significantly outperforms all baselines. Additionally, we observe consistent improvements over BA-TFD and BA-TFD+ \cite{cai2022you}, outperforming them by a wide margin on both datasets. On AV-Deepfake1M++, we achieve a video-level \textbf{AUC of 96.66\%}, surpassing the audio-visual baseline UMMAFormer (87.04\%) by a large margin (+9.6\%). Notably, our method achieves an exceptional \textbf{AUC (Seg) of 98.23\%}, demonstrating that our multi-branch architecture effectively captures forgery traces even in short 320 ms segments where other methods struggle.

\smallskip
\noindent \textbf{Cross-Dataset Generalization.}
In the cross-dataset evaluation on FakeAVCeleb, our method achieves an AUC of \textbf{83.03\%}, significantly outperforming MM-Det (76.74\%) and UMMAFormer (62.29\%). This indicates that our LMM-guided semantic tokens and multi-scale audio features learn generalized forgery representations rather than overfitting to specific dataset artifacts.

\smallskip
\noindent \textbf{Ablation Study.}
We further validate the significance of components from our framework in Table \ref{tab:ablation}. Notably, removing the semantic branch (Variant 0) causes a severe performance drop compared to the text-visual baseline (Variant 1), confirming the crucial role of LMM-derived features. Adding the audio branch (Variant 2) significantly improves performance over the text-visual baseline. Furthermore, the inclusion of the OneCycleLR scheduler (Ours) provides the final boost necessary to achieve state-of-the-art results, confirming the importance of stable optimization in multi-modal training.

% --- TABLE: ABLATION (Placed immediately after Detection table) ---

\begin{table}[t]
\vspace{-3mm}
\centering
\caption{\small \textit{Ablation study on model and training strategies (\%)}.}
\label{tab:ablation}
\resizebox{\linewidth}{!}{
\begin{tabular}{c|ccc|c|cc}
\hline
\textbf{Method} & \textbf{Text} & \textbf{Visual} & \textbf{Audio} & \textbf{OneCycleLR} & \textbf{AUC (Video)}$\uparrow$ & \textbf{AUC (Seg)}$\uparrow$ \\ \hline
Variant 0 & - & \checkmark & - & - & 56.61 & 69.11 \\
Variant 1 & \checkmark & \checkmark & - & - & 85.44 & 83.78 \\
Variant 2 & \checkmark & \checkmark & \checkmark & - & 96.21 & 97.04 \\
\textbf{Ours} & \checkmark & \checkmark & \checkmark & \checkmark & \textbf{96.66} & \textbf{98.23} \\ \hline
\end{tabular}
}
\vspace{-6mm}
\end{table}

% ==========================================
% LOCALIZATION PERFORMANCE
% ==========================================

\smallskip
\noindent \textbf{Localization Performance.}
% --- TABLE: LOCALIZATION (Placed strictly at start of Localization section) ---
\begin{table}[t]
\centering
\caption{\small \textit{Comparison of temporal localization performance (\%)}.}
\label{tab:localization}
\resizebox{\linewidth}{!}{
\begin{tabular}{c|cccc|ccccc}
\hline
\multirow{2}{*}{\textbf{Method}} & \multicolumn{4}{c|}{\textbf{Average Precision (AP)} $\uparrow$} & \multicolumn{5}{c}{\textbf{Average Recall (AR)} $\uparrow$} \\
 & \textbf{@0.5} & \textbf{@0.75} & \textbf{@0.9} & \textbf{@0.95} & \textbf{@50} & \textbf{@30} & \textbf{@20} & \textbf{@10} & \textbf{@5} \\ \hline
BA-TFD & 14.10 & 0.67 & 0.01 & 0.00 & 23.21 & 19.01 & 14.23 & 11.42 & 10.35 \\
BA-TFD+ & 18.12 & 7.13 & 0.64 & 0.01 & 25.12 & 21.01 & 19.22 & 18.41 & 15.32 \\
% \hline
PartialSpoof (Audio) & 22.12 & 2.02 & 0.13 & 0.00 & 15.07 & 15.06 & 15.04 & 15.01 & 14.97 \\
Modified MM-Det & 2.32 & 0.09 & 0.04 & 0.01 & 3.99 & 3.97 & 3.95 & 3.84 & 3.49 \\ \hline
\textbf{Ours} & \textbf{52.89} & \textbf{12.67} & \textbf{0.78} & \textbf{0.06} & \textbf{40.86} & \textbf{40.77} & \textbf{40.76} & \textbf{40.72} & \textbf{39.74} \\ \hline
\end{tabular}
}
\vspace{-0.5cm}
\end{table}
To fairly evaluate the fine-grained temporal localization capabilities of our framework, we compare our method against state-of-the-art localization methods and constituent modality baselines. Specifically, we compare against: (1) \textbf{BA-TFD} and \textbf{BA-TFD+} \cite{cai2022you}, (2) \textbf{PartialSpoof}, representing the audio-only localization performance, and (3) \textbf{Modified MM-Det}, an adaptation of \cite{song2025learningmultimodalforgeryrepresentation} using our sliding-window logic. As shown in Table \ref{tab:localization}, our method significantly outperforms all baselines, surpassing BA-TFD+ by roughly \textbf{+34\%} in terms of AP@0.5.

% ==========================================
% QUALITATIVE ANALYSIS
% ==========================================
\vspace{-3mm}
\subsection{Qualitative Analysis}
\vspace{-2mm}
To validate our multi-branch fusion strategy, we visualize the frame-level detection scores across four representative scenarios in Fig. \ref{fig:qualitative}. These scenarios range from fully authentic clips to ``partial" forgeries, where only the audio or the visual track is manipulated.

A key challenge in deepfake detection is handling these partial forgeries. For instance, in the \textit{Real Visual Fake Audio} case (Fig. \ref{fig:qualitative}, third subfigure), the visual content is authentic. Consequently, the Visual branch (green line) outputs a near-zero score, effectively missing the forgery. However, the Audio branches (blue/purple lines) successfully detect the spoofing artifacts and spike in value.

Crucially, our Fusion module (red line) robustly handles this discrepancy. By aggregating the signals, the Fusion module allows the strong audio detection to override the lack of visual evidence, resulting in a high detection score. This confirms that our framework can accurately localize forgeries even when evidence is present in only one modality.

\vspace{-5mm}
\section{CONCLUSION}
\vspace{-3mm}
In this paper, we proposed a comprehensive three-branch framework for AI-generated video detection and localization that jointly models visual artifacts, audio spoofing cues, and semantic signals. By integrating a Large Multi-modal Model to capture high-level semantic inconsistencies alongside dedicated spatio-temporal and audio branches, we address the limitations of existing binary classification methods. Extensive experiments on large-scale benchmarks demonstrate that our approach not only achieves state-of-the-art detection performance but also effectively performs fine-grained temporal localization of partially manipulated segments.

\noindent \textbf{Limitations.} One limitation of our proposed framework is that it is a two-stage approach, as the visual and audio branches are currently trained separately before fusion.
% Despite its effectiveness, our current pipeline exhibits certain limitations. First, the preprocessing stage is computationally intensive and storage-heavy; specifically, the ST and LMM branches require frame extraction and intermediate reconstruction via VQ-VAE, making the data preparation process tedious. Second, the training process is cumbersome as the visual and audio branches are currently trained separately before fusion. Third, the model faces scalability challenges with long videos (typically exceeding 40 seconds) due to the high memory complexity required for maintaining dense, time-aligned token sequences for localization. Finally, while we achieved high metrics on specific datasets like AV-Deepfake1M++, the diversity of real-world attacks remains a challenge, and further evaluation on broader datasets is necessary to identify potential corner cases.

\noindent \textbf{Future Work.} We plan to streamline the pipeline by exploring end-to-end training strategies that eliminate the need for disjoint optimization of separate branches. We also plan to investigate the construction of a unified audio-visual latent space to allow the modalities to learn joint representations more efficiently, potentially reducing the model's complexity. 
% Additionally, we will focus on optimizing the memory overhead to handle longer video sequences and validate the framework's robustness against a wider array of unseen generative techniques.
% \vspace{-9mm}
% \section*{Acknowledgments} 
% \vspace{-4mm}

\noindent \textbf{Acknowledgments.}This work is supported by the U.S. National Science Foundation (NSF) under grant IIS-2434967, the National Artificial Intelligence Research Resource (NAIRR) Pilot and TACC, and Purdue Applied AI Research Center.
The views, opinions, and/or findings expressed are those of the author and should not be interpreted as representing the official views of NSF, NAIRR Pilot, and Purdue University.

\bibliographystyle{IEEEbib}
% \bibliography{strings,refs}
\bibliography{refs}

\begin{thebibliography}{10}

\bibitem{team2023gemini}
Gemini Team, Rohan Anil, Sebastian Borgeaud, Jean-Baptiste Alayrac, Jiahui Yu, Radu Soricut, Johan Schalkwyk, Andrew~M Dai, Anja Hauth, and Katie Millican,
\newblock ``Gemini: a family of highly capable multimodal models,''
\newblock {\em arXiv preprint arXiv:2312.11805}, 2023.

\bibitem{lin2024detecting}
Li~Lin, Neeraj Gupta, Yue Zhang, Hainan Ren, Chun-Hao Liu, Feng Ding, Xin Wang, Xin Li, Luisa Verdoliva, and Shu Hu,
\newblock ``Detecting multimedia generated by large ai models: A survey,''
\newblock {\em arXiv preprint arXiv:2402.00045}, 2024.

\bibitem{lin2024preserving}
Li~Lin, Xinan He, Yan Ju, Xin Wang, Feng Ding, and Shu Hu,
\newblock ``Preserving fairness generalization in deepfake detection,''
\newblock in {\em Proceedings of the IEEE/CVF conference on computer vision and pattern recognition}, 2024, pp. 16815--16825.

\bibitem{Zhang_2023}
Rui Zhang, Hongxia Wang, Mingshan Du, Hanqing Liu, Yang Zhou, and Qiang Zeng,
\newblock ``Ummaformer: A universal multimodal-adaptive transformer framework for temporal forgery localization,''
\newblock in {\em Proceedings of the 31st ACM International Conference on Multimedia}. Oct. 2023, MM ’23, p. 8749–8759, ACM.

\bibitem{song2025learningmultimodalforgeryrepresentation}
Xiufeng Song, Xiao Guo, Jiache Zhang, Qirui Li, Lei Bai, Xiaoming Liu, Guangtao Zhai, and Xiaohong Liu,
\newblock ``On learning multi-modal forgery representation for diffusion generated video detection,''
\newblock {\em Advances in Neural Information Processing Systems}, vol. 37, pp. 122054--122077, 2024.

\bibitem{liu2023visualinstructiontuning}
Haotian Liu, Chunyuan Li, Qingyang Wu, and Yong~Jae Lee,
\newblock ``Visual instruction tuning,''
\newblock {\em Advances in neural information processing systems}, vol. 36, pp. 34892--34916, 2023.

\bibitem{radford2021learningtransferablevisualmodels}
Alec Radford, Jong~Wook Kim, Chris Hallacy, Aditya Ramesh, Gabriel Goh, Sandhini Agarwal, Girish Sastry, Amanda Askell, Pamela Mishkin, and Jack Clark,
\newblock ``Learning transferable visual models from natural language supervision,''
\newblock in {\em International conference on machine learning}. PmLR, 2021, pp. 8748--8763.

\bibitem{van2017neural}
Aaron Van Den~Oord and Oriol Vinyals,
\newblock ``Neural discrete representation learning,''
\newblock {\em Advances in neural information processing systems}, vol. 30, 2017.

\bibitem{10003971}
Lin Zhang, Xin Wang, Erica Cooper, Nicholas Evans, and Junichi Yamagishi,
\newblock ``The partialspoof database and countermeasures for the detection of short fake speech segments embedded in an utterance,''
\newblock {\em IEEE/ACM Transactions on Audio, Speech, and Language Processing}, vol. 31, pp. 813--825, 2023.

\bibitem{NEURIPS2020_92d1e1eb}
Alexei Baevski, Yuhao Zhou, Abdelrahman Mohamed, and Michael Auli,
\newblock ``wav2vec 2.0: A framework for self-supervised learning of speech representations,''
\newblock in {\em Advances in Neural Information Processing Systems}, H.~Larochelle, M.~Ranzato, R.~Hadsell, M.F. Balcan, and H.~Lin, Eds. 2020, vol.~33, pp. 12449--12460, Curran Associates, Inc.

\bibitem{DBLP:journals/corr/abs-2105-08050}
Hanxiao Liu, Zihang Dai, David~R. So, and Quoc~V. Le,
\newblock ``Pay attention to mlps,''
\newblock {\em CoRR}, vol. abs/2105.08050, 2021.

\bibitem{cai2025av}
Zhixi Cai, Kartik Kuckreja, Shreya Ghosh, Akanksha Chuchra, Muhammad~Haris Khan, Usman Tariq, Tom Gedeon, and Abhinav Dhall,
\newblock ``Av-deepfake1m++: A large-scale audio-visual deepfake benchmark with real-world perturbations,''
\newblock in {\em Proceedings of the 33rd ACM International Conference on Multimedia}, 2025, pp. 13686--13691.

\bibitem{wang2021comparativestudyrecentneural}
Xin Wang and Junich Yamagishi,
\newblock ``A comparative study on recent neural spoofing countermeasures for synthetic speech detection,''
\newblock {\em arXiv preprint arXiv:2103.11326}, 2021.

\bibitem{DBLP:journals/corr/abs-1905-02479}
Xiao Zhang, Rui Zhao, Junjie Yan, Mengya Gao, Yu~Qiao, Xiaogang Wang, and Hongsheng Li,
\newblock ``P2sgrad: Refined gradients for optimizing deep face models,''
\newblock {\em CoRR}, vol. abs/1905.02479, 2019.

\bibitem{smith2019super}
Leslie~N Smith and Nicholay Topin,
\newblock ``Super-convergence: Very fast training of neural networks using large learning rates,''
\newblock in {\em Artificial intelligence and machine learning for multi-domain operations applications}. SPIE, 2019, vol. 11006, pp. 369--386.

\bibitem{khalid2022fakeavcelebnovelaudiovideomultimodal}
Hasam Khalid, Shahroz Tariq, Minha Kim, and Simon~S Woo,
\newblock ``Fakeavceleb: A novel audio-video multimodal deepfake dataset,''
\newblock {\em Advances in Neural Information Processing Systems}, 2021.

\bibitem{cai2022you}
Zhixi Cai, Kalin Stefanov, Abhinav Dhall, and Munawar Hayat,
\newblock ``Do you really mean that? content driven audio-visual deepfake dataset and multimodal method for temporal forgery localization,''
\newblock in {\em 2022 International Conference on Digital Image Computing: Techniques and Applications (DICTA)}, Sydney, Australia, 2022, pp. 1--10.

\end{thebibliography}

\end{document}